\pdfoutput=1
\documentclass[letterpaper, 10 pt, conference]{ieeeconf}
\IEEEoverridecommandlockouts                              

\usepackage{multicol}

\usepackage{graphics} 
\usepackage{epsfig} 
\usepackage{amsmath} 
\usepackage{amssymb}
\usepackage[dvipsnames]{xcolor}
\definecolor{peach}{RGB}{255,220,102}

\usepackage[percent]{overpic}

\usepackage{transparent}

\usepackage{booktabs}
\newcommand{\ra}[1]{\renewcommand{\arraystretch}{#1}}
\usepackage{caption, copyrightbox}
\newcommand{\keypoint}[1]{\vspace{0.2cm}\noindent\textit{#1}\quad}
 
\newcommand{\cut}[1]{}
\newcommand{\changed}[1]{\textcolor{black}{#1}}

\title{\LARGE \bf
Adversarial Generation of Informative Trajectories for Dynamics System Identification
}

\author{Marija Jegorova, Joshua Smith, Michael Mistry, Timothy Hospedales$^{1,2}$
\thanks{$^{1}$ School of Informatics, University of Edinburgh, UK. \texttt{m.jegorova@ed.ac.uk, joshua.smith@ed.ac.uk, m.mistry@ed.ac.uk, t.hospedales@ed.ac.uk}.
$^2$ This work is supported by EPSRC under Grant EP/R026173/1.}}

\begin{document}

\maketitle{}
\thispagestyle{empty}
\pagestyle{empty}

\begin{abstract}

Dynamic System Identification approaches usually heavily rely on evolutionary and gradient-based optimisation techniques to produce optimal excitation trajectories for determining the physical parameters of robot platforms. Current optimisation techniques tend to generate single trajectories. This is expensive, and intractable for longer trajectories, thus limiting their efficacy for system identification. We propose to tackle this issue by using multiple shorter cyclic trajectories, which can be generated in parallel, and subsequently combined together to achieve the same effect as a longer trajectory. 
Crucially, we show how to scale this approach even further by increasing the generation speed and quality of the dataset through the use of generative adversarial network (GAN) based architectures to produce large databases of valid and diverse excitation trajectories.
To the best of our knowledge, this is the first robotics work to explore system identification with multiple cyclic trajectories and to develop GAN-based techniques for scaleably producing excitation trajectories that are diverse in both control parameter and inertial parameter spaces.
We show that our approach dramatically accelerates trajectory optimisation, while simultaneously providing more accurate system identification than the conventional approach.
\end{abstract}
\vspace*{-1 pt}
\setlength{\belowcaptionskip}{-10pt}
\section{Introduction}
In the light of the continuous improvement in robotic mechanical design, the importance of accurate models for robot dynamics increases immensely. Model inaccuracies can have a significant effect on control, stability, and motion optimisation of the platforms. Hence the problem of system identification in robotics is currently being revisited.
All dynamics system identification techniques are highly data-dependent in the sense that their parameter estimation efficacy, and the generalisation of dynamics models using these parameters, are highly affected by the quality of the trajectories used for exploration.

Most conventional system identification methods rely on a single parameterised trajectory \cite{swevers1997,Park2006,bethge2017}. Such trajectories are limited in how much they can explore system parameters  within their set length. Extending the length of this trajectory alleviates this limitation, however the computation required to generate an optimal excitation trajectory grows rapidly with trajectory length, and quickly becomes intractable. 
We explore whether generating multiple shorter diverse trajectories can be used to achieve the same effect more scaleably than existing methods -- or to outperform them -- by effectively allowing the generation of longer overall trajectories. 
If many shorter trajectories, diverse in the inertial parameter space, can be generated, they can ultimately better explore all the parameters than a single longer trajectory, while being easy and cheap to generate in parallel. Furthermore, assuming that they are cyclic (same start and end condition), the set of short trajectories can be concatenated for easy execution in sequence. We provide experimental results to confirm that generating multiple  shorter excitation trajectories tends to be better than the equivalently long single excitation trajectory in terms of system identification performance.


To fully leverage this paradigm of system identification trajectory generation, we need the ability to efficiently generate numerous diverse short excitation trajectories. To this end, we propose a pipeline where a traditional trajectory optimizer is used to generate an initial seed dataset, after which we train a Generative Adversarial Network (GAN) on this seed set. Once trained the GAN provides a surrogate model for excitation trajectory optimisation that can effectively generate an unlimited number of diverse short excitation trajectories rapidly \changed{(roughly four orders of magnitude faster than the conventional approach)} and in parallel. 
Our generative model is optimised with respect to the validity (in terms of the constraints and avoiding self-collisions) and fitness (excitation) scores, in order to generate a dataset that is maximally informative about system dynamics. 

Our proposed method is called System IDEntification GAN (SIDE-GAN). In principle, the SIDE-GAN is indifferent to the dynamics model behind the training trajectories. So, given the suitable initial training dataset, SIDE-GAN can provide quality training data for either parametric, semi-parametric, or even for some non-parametric system models. Our empirical results show that our SIDE-GAN approach improves the accuracy of parameter estimation and torque prediction, with a recursive least squares identified model. Furthermore, our generation speed increases by orders of magnitude compared to the original short trajectories optimisation, and allow us to generate a total excitation trajectory length that is intractable with traditional long-trajectory optimisation. 
Our empirical results are demonstrated using both real and simulated  KUKA LWR IV manipulation platforms.
\vspace{-10pt}
\section{Related Work}
\vspace{-10pt}
\keypoint{Current models for dynamics system identification.} Previous work on generating exciting trajectories for inertial parameter identification has mainly focused on approaches that optimize a single parameterized trajectory for optimal excitation \cite{swevers1997,Park2006,bethge2017}. Typically, this is achieved by maximizing the identifiability of each parameter
via the stacked regressor matrix  from the trajectory that is being optimised (please refer to the Section~\ref{sec:motivation} for more details). 

The optimization metrics tend to be highly non-linear with even more complex constraints, such as avoiding self-collision or certain regions of space completely. The optimization task is thus extremely difficult due to non-smoothness and many local minima. Genetic algorithms \cite{bethge2017} are often used, but take many iterations to converge. Moreover, the cost of evaluating a single step of this optimizations is more than $O(n^3)$ in the desired length of the trajectory.

To our knowledge no previous work has explored optimizing multiple trajectories. Our divide-and-conquer approach can generate longer (and thus more informative) trajectories than the traditional approach. Besides being over two orders of magnitude faster, we find that our multi-trajectory can outperform the standard approach, even when controlling for the total length of the final trajectory. 




\keypoint{Generative Adversarial Networks} GANs \cite{vanillaGAN} are a family of neural network methods, that have gained popularity for realistic image   generation since 2014. They have since been applied to a multitude of tasks, although their primarily focus has largely remained realistic image and video generation \cite{LargeScaleGAN, SuperGAN, text2image, videoGAN, vid2vid}, for example from captions. 

Compared to the image generation area, there is still comparatively little research on how GANs can be of significant help in the field of robotics. Conventional image GANs can generate data to assist training visual recognition systems in autonomous systems~\cite{ICRAsonar}, however applications of GANs to planning and control are still very sparse.

The most relevant GAN-based approaches to non-sensory part of robotics so far include imitation learning~\cite{GAIL, infoGAIL} -- to efficiently learn a single policy or a discrete set of policies from demonstration; and direct generation of robot control policy repertoires~\cite{GPN}. The latter provides robust goal-directed control by enabling sampling diverse controllers from a continuous goal-conditional distribution over control policies. 
There has also been some research conducted on learning the inverse kinematics (IK) of the robot using GAN-like architectures~\cite{IK_GAN}. However this work does not leverage the diversity potential of GANs, discarding the random noise input \cut{$z\in{Z}$} completely, instead replacing it with the end-effector position (for the IK problem). This replacement strips the generator of all the diverse generative properties, boiling it down to merely a mapping network, and the whole architecture to an actor-critic-like model.

We provide the first application of GAN-like methods to the `experimental design' aspect of dynamics system identification. We learn our SIDE-GAN that on a seed set of excitation trajectories, which then provides a surrogate model to replace the typical compute intensive trajectory optimisation process. We rely on the ability of the trained GAN to rapidly generate \emph{diverse} cyclic trajectories which can then be combined to provide an informative long trajectory.



\vspace{6pt}
\section{Problem and Motivation}\label{sec:motivation}
\keypoint{Dynamics system identification} is the task of learning the inertial parameters $\pi$ of the links of the robot. Normally this is achieved by starting with the Rigid Body Dynamics (RBD) equation:
\begin{equation}
    \tau = \mathbf{M}(q)\ddot{q} + \mathbf{C}(q, \dot{q})\dot{q} + G(q) + F(q,\dot{q})
\end{equation}
Where $\mathbf{M}$ represents the inertia matrix, $\mathbf{C}$ the Coriolis and centrifugal matrix, $G$ is the gravity vector, $F$ is the friction vector and $\tau$ is the full joint torques experienced and measured by the robotic platform. The state of the robot is expressed in terms of $(q,\dot{q},\ddot{q})$ - position, velocity, and acceleration.
We then use the RBD equation and rearrange it \changed{with a regressor $Y(q,\dot{q},\ddot{q})$} to form Eq.~\eqref{eqn:RBDRearranged}, which is linear with respect to $\pi$. This allows us to the use standard least squares approach \cite{Atkeson1986} to solve for $\pi$, shown in \eqref{eqn:LSSingle}.
\begin{multicols}{2}\noindent
\begin{equation}
    \tau = Y(q,\dot{q},\ddot{q})\pi\label{eqn:RBDRearranged}
\end{equation}
\begin{equation}
     Y^{-1}(q,\dot{q},\ddot{q})\tau = \pi\label{eqn:LSSingle}
\end{equation}
\end{multicols}
This solution would be valid for a single state of the robot but would rarely be the correct model due to noise in the data. Typically, the dynamics is sampled at many different input states (positions, velocities, and accelerations) to compensate for the noise. This allows multiple regressors to be constructed in a stacked matrix, alongside their equivalent stacked torque vector:
\begin{multicols}{2}\noindent
\begin{equation}
     Y = \begin{bmatrix} Y(q_0,\dot{q}_0,\ddot{q}_0) \\...\\ Y(q_n,\dot{q}_n,\ddot{q}_n)\end{bmatrix}
\end{equation}
\begin{equation}
     \tau_{s} = \begin{bmatrix} \tau_0 \\ ... \\\tau_n\end{bmatrix}
\end{equation}
\end{multicols}

We can then estimate the inertial parameters $\pi$ using the pseudo-inverse of the regressor matrix:
\begin{equation}
    Y^{T}(YY^{T})^{-1} \tau_s = \pi\label{eqn:stackRegressor}
\end{equation}
Numerical errors can occur if $YY^{T}$ is ill-conditioned (e.g., has very small or zero eigenvalues). When regressors are ill-conditioned, the trajectories sampled to provide  state-torque pairs do not sufficiently excite the relevant parameters. For example, a sampled trajectory may not accelerate enough for the inertia to affect the output torques. With low excitation these parameters are not identifiable from the sampled data, and the regressor will be ill-conditioned.

\keypoint{Discussion: Non-identifiable parameters} Please note that some parameters are always non-identifiable as they have no effect on the output torque no matter the state of the robot. These non-identifiable parameters can be removed by calculating the base parameter set of the robot \cite{Gautier1991} which will let us replace the regressor matrices with a base regressor matrix, $Y_b$, and replace the inertial parameter with the base inertial parameters, $\pi_b$, which contain the set of parameters that are excitable, such that \eqref{eqn:baseDynamics} holds. From this point forward when the stacked regressor $Y$ is referred to, it contains the base regressors of each state rather than the full regressor, and with $\pi$ replaced by $\pi_b$.
\begin{equation}
    \tau = Y_b(q,\dot{q},\ddot{q}) \pi_b \label{eqn:baseDynamics}
\end{equation}

\begin{figure*}
    \centering
    \vspace{0.1cm}
    \includegraphics[scale=0.3]{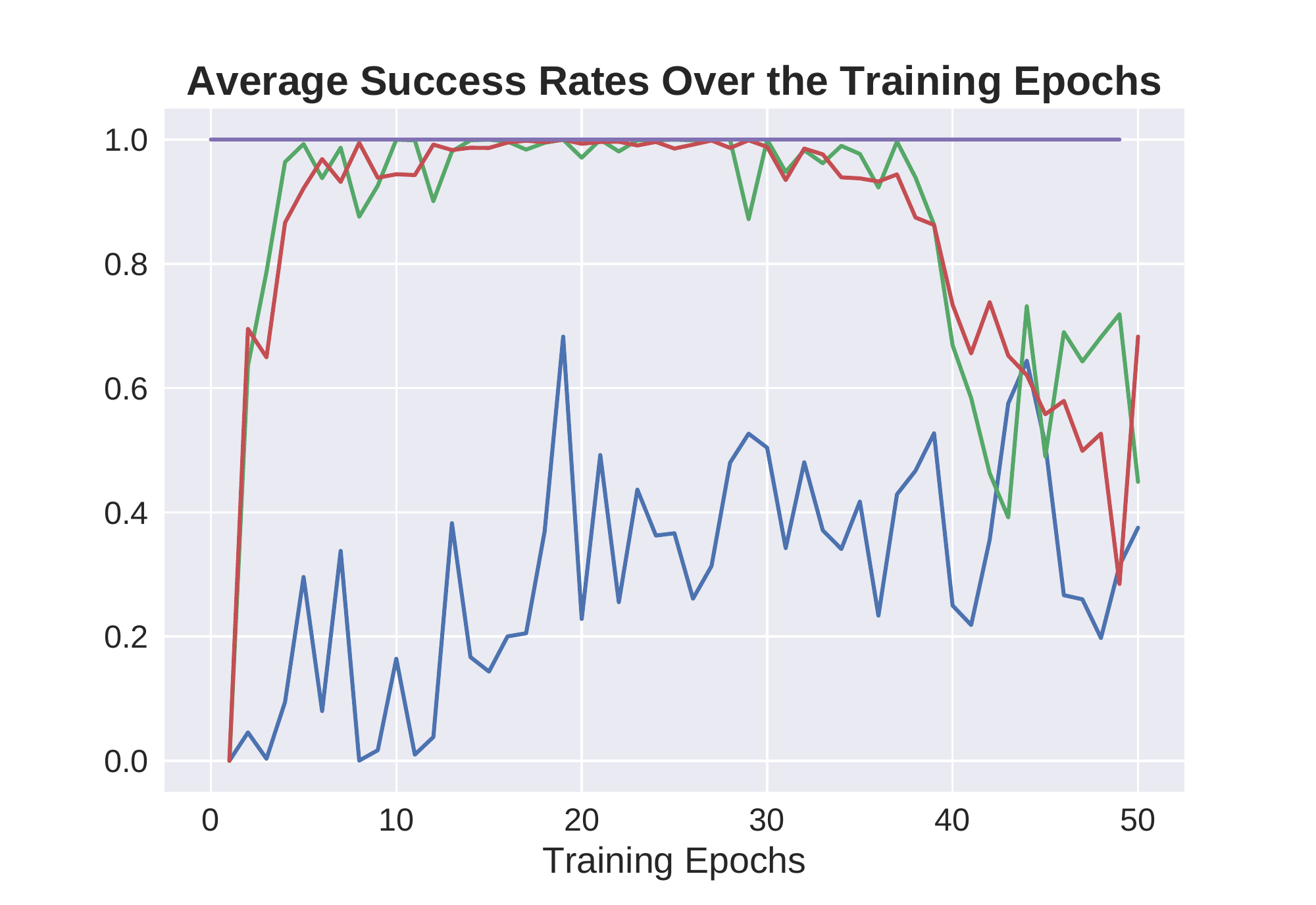}\includegraphics[scale=0.3]{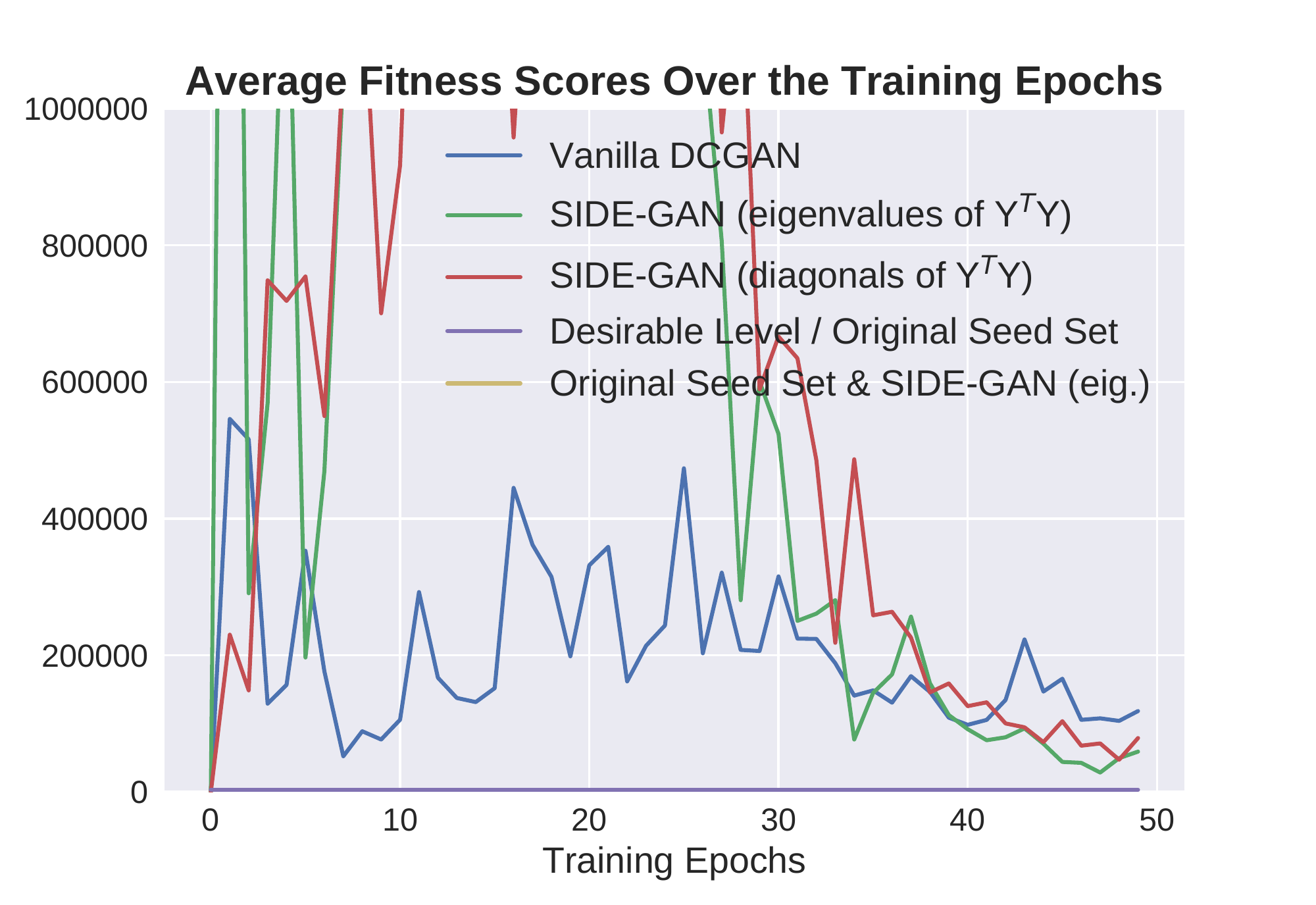}
    \begin{minipage}[b][113pt][s]{155pt}
      \centering
      \includegraphics[height=50pt,width=165pt]{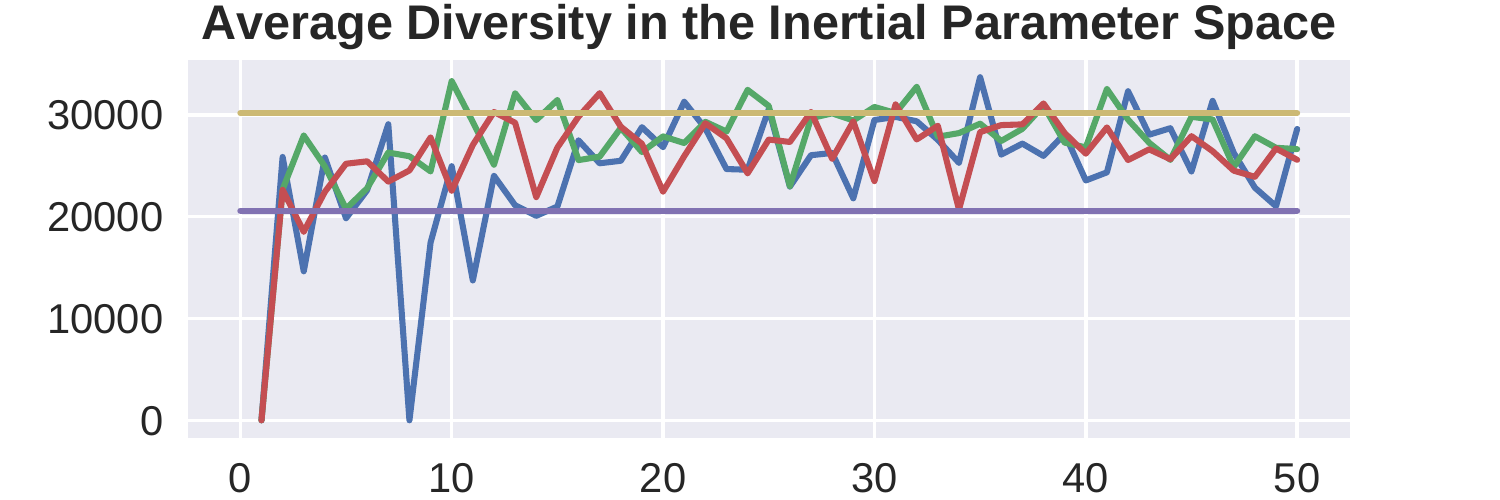}
      \includegraphics[height=62pt,width=142pt]{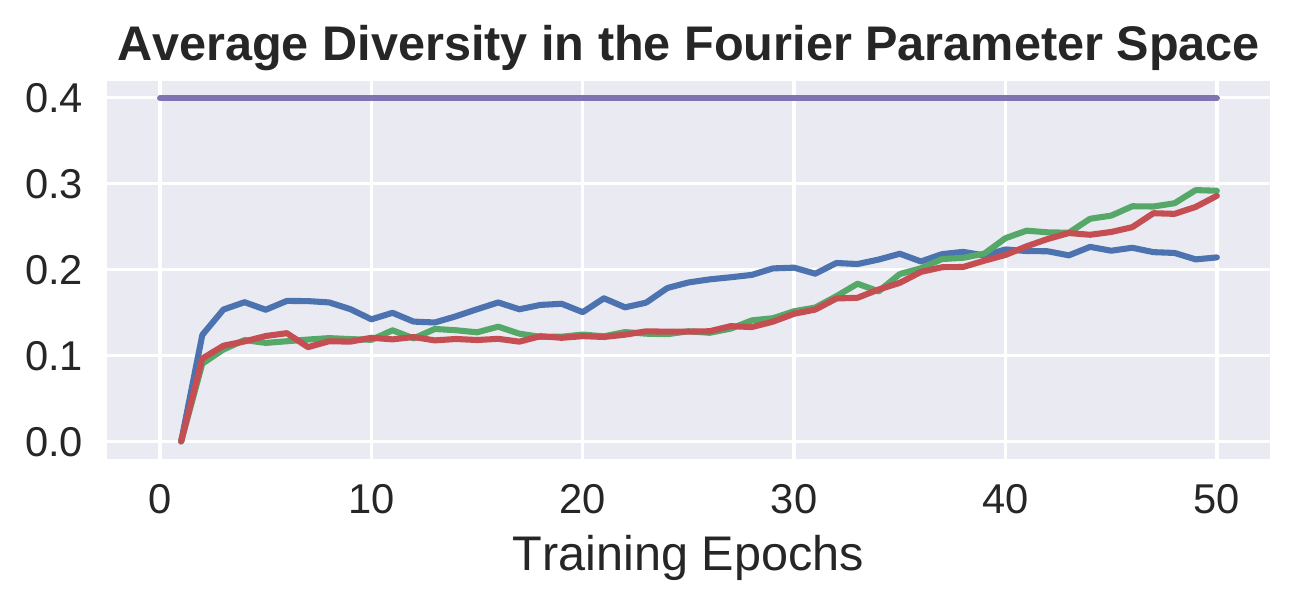}
    \end{minipage}
    \vspace{-5pt}
    \caption{\textbf{Average metrics over the SIDE-GAN training time:} the SIDE-GAN is trained for 50 epochs in total, the averages are taken over 3000 trajectories per epoch for each of the methods, i.e., 3 training runs, 10 noise vectors, 100 trajectories (batch size) produced by each input noise vector. \textbf{1.} Average success rate - i.e. ratio of the valid trajectories SIDE-GAN generates on average. \textbf{2.} Average fitness - i.e., the average condition number of the $Y^{T}Y$ for the entire trajectory. The lower the fitness score, the better is the quality of the trajectory for the system identification purposes. \changed{\textbf{1} and \textbf{2} represent the SIDE-GAN training\cut{ constraints}. SIDE-GAN learns to generate successful (valid) trajectories within $\leq10$ training epochs, however better (lower) fitness scores is harder to achieve, SIDE-GANs trade it off for lower validity scores at training epochs 30-50.} \textbf{3.} The diversity of the generated batch is assessed via the average pairwise Euclidean distance between trajectories in two spaces of interest. The bottom: diversity in the Fourier parameter space (immediate output of the generator) - at epochs 40-50 the SIDE-GAN overtakes vanilla DCGANs. \changed{The gap between the desirable level vs. SIDE-GANs shows that synthetic data have comparable but still worse quality than real.} The top: the average pairwise Euclidean distance between $\text{diag}(Y^{T}Y)$ of the trajectories in the generated batches, representing the batch-diversity in the inertial parameters space. The SIDE-GANs diversity is usually equal or higher than that of the vanilla DCGANs, significantly surpassing the original seed set diversity (purple). The average inertial parameter diversity of the SIDE-GAN data (generated after 50 training epochs) and original training set put together (yellow) shows that adding the SIDE-GAN data to the original dataset is highly beneficial.}
    \label{figure:gan_training} 
    \vspace{-5pt}
\end{figure*}

\keypoint{Quantifying Trajectory Excitation}
As discussed earlier, unexciting trajectories can lead to base regressors that are still low rank with low excitation, meaning ill-conditioned $YY^T$  and numerical errors in Eq.~\eqref{eqn:stackRegressor}. The goal of trajectory optimization is to produce trajectories $((q_0,\dot{q}_0,\ddot{q}_0,\tau_0),\dots,(q_n,\dot{q}_n,\ddot{q}_n,\tau_n))$ that lead to well-conditioned $YY^T$ and accurate estimation of parameters $\pi$. There are two different objectives correlated with the trajectory quality, that can be used during the SIDE-GAN training. In section~\ref{sec:exp2} we show that either can be used for training the SIDE-GAN, and results are comparable:


\noindent\textbf{Eigenvalue Fitness}: The first metric is condition number, i.e., minimal ratio between the largest and smallest eigenvalues of $Y^{T}Y$. This implies a high parameter excitement {\textit{within the trajectory}, with the least and the most excited inertial parameters being explored as equally as possible,} directly leading to a better conditioned $YY^{T}$ matrix. We refer to the condition number of $Y^{T}Y$ as `fitness', and use it to report the progress of training in Figure~\ref{figure:gan_training}.

\noindent \textbf{Diagonal Fitness}: 
Another objective is based on the trace of $Y^{T}Y$, i.e. either maximising the trace itself or minimizing the trace of the inverse. The diagonal of $Y^{T}Y$ is indicative of the level of exploration trajectory does in the inertial parameter space. The trace is indifferent to basis changes and hence is more comparable \textit{across the trajectories} with different basis vectors.  
We refer to
\begin{equation}
    \dfrac{\textrm{max}(\text{diag}(Y^{T}Y))}{\textrm{min}(\text{diag}(Y^{T}Y))}\label{eq:diagFitness}
\end{equation}
 as the `diagonal fitness', and use it for training one of the two versions of the SIDE-GAN assessed in Section~\ref{sec:exp2}. Similarly to the conventional fitness, smaller scores correspond to more evenly explored inertial parameters.

\keypoint{System Identification} The conventional approach to dynamics system identification optimizes for a trajectory with high fitness, executes this trajectory on the robot, and then estimates inertial parameters as in Eq.~\ref{eqn:stackRegressor}. However, given the cost of optimizing long trajectories, the achievable fitness and parameter identification accuracy is limited.

\keypoint{Use of Multiple Trajectories} The unique aspect of our approach is to use multiple trajectories to improve inertial parameter exploration. For the  purpose of system identification, we desire not only diversity in control parameter space, but also to explore different subsets of inertial parameters. Then their stacked regressors combined will 
have more uniformly distributed eigenvalues, and when inverted will produce a better estimate of inertial parameters $\pi$.
The eigenvalues of the stack of short trajectories correspond to inertial parameter identifiability, as for the conventional single-trajectory approach. We use Modified Fourier Trajectories \cite{Park2006}, \cut{which are }cyclic in the start and end conditions, making concatenation of several trajectories together trivial.  We investigate: Can a sufficiently large and diverse set of cyclic trajectories be generated? And: How such set of trajectories performs for system identification compared to a standard single longer trajectory?

Using multiple short trajectories also has an important advantage over a single long one in terms of generation efficiency. The time of optimizing the trajectories grows quickly as the trajectories grow longer, due to the $O(n^3)$ complexity computing $Y^{T}Y$ and the additional complexity of computing the self-collisions at each discrete step. Optimising multiple short trajectories means much faster regressor calculation, as well as enabling trivial parallelisation.


\keypoint{Summary} Our proposed pipeline uses conventional optimisation to generate a set of short trajectories, then trains SIDE-GAN to to rapidly expand this dataset. 
The ultimate testing objective is then to show that model-based torque prediction (using parameters obtained from system identification) is more accurate when using the SIDE-GAN-generated trajectories than based on just the initial seed data, or the conventional single longer trajectory.

\begin{figure}
    \centering
    \vspace{0.2cm}
    \includegraphics[scale=1]{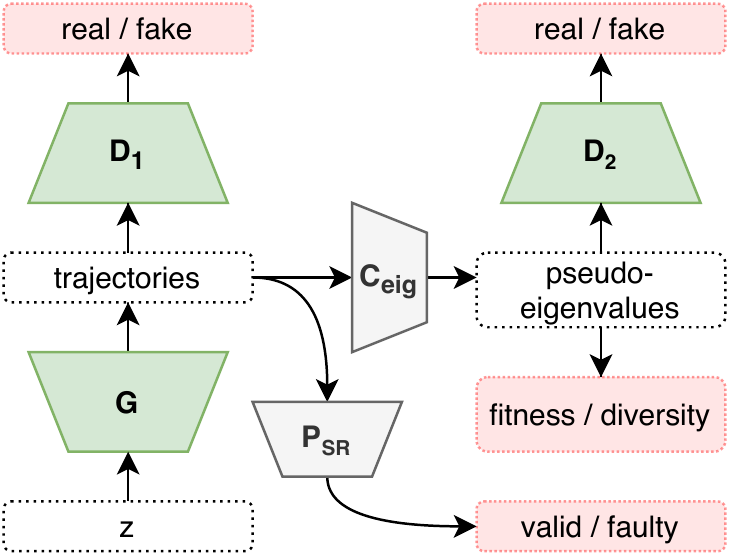}
    \caption{\textbf{SIDE-GANs at training time:} In the conventional GAN architecture on the left, the generator $G$ inputs a \changed{uniform random} noise vector $z$, and outputs synthetic trajectories \changed{(Fourier transform parameters)}. The discriminator $D_1$ tries to distinguish ``fake'' (synthetic) vs real trajectories.
    \\
    The right side of the scheme represents the new part of the system, where the pre-trained predictor $P_{SR}$ predicts valid/faulty, and thus provides a penalty for invalid trajectories.  The pre-trained converter $C_{eig}$ maps trajectories to their estimated eigenvalues. These pseudo-eigenvalues are then assessed by the second discriminator $D_2$ as  ``real'' or ``fake'', they are also used as input to compute an eigenvalue fitness penalty that encourages high excitation trajectories.  
    \\
    \textbf{Colour-coding:} Trapezoid blocks are neural networks. Grey trapezoids correspond to pre-trained networks with weights frozen for the main SIDE-GAN system training. Green trapezoids correspond to the networks that are learned during main system training. \cut{Pink blocks are used during training to compute losses.}\changed{Pink blocks stand for training losses.}
    }\vspace{-5pt}
    \label{figure:scheme} 
\end{figure}

\vspace{0pt}
\section{Method and Architecture}\label{method}
\vspace{-5pt}
\keypoint{Architecture} 
SIDE-GAN is built for the generation of diverse excitation trajectories for system identification.  Figure~\ref{figure:scheme} shows the  architecture. SIDE-GAN is trained on a set of seed trajectories generated by the conventional optimizer, and once trained can rapidly generate new trajectory batches. 

\begin{figure*}
    \centering
    \vspace{0.2cm}
    \begin{minipage}{1.03\columnwidth}
      \centering
      \includegraphics[width=0.2325\textwidth]{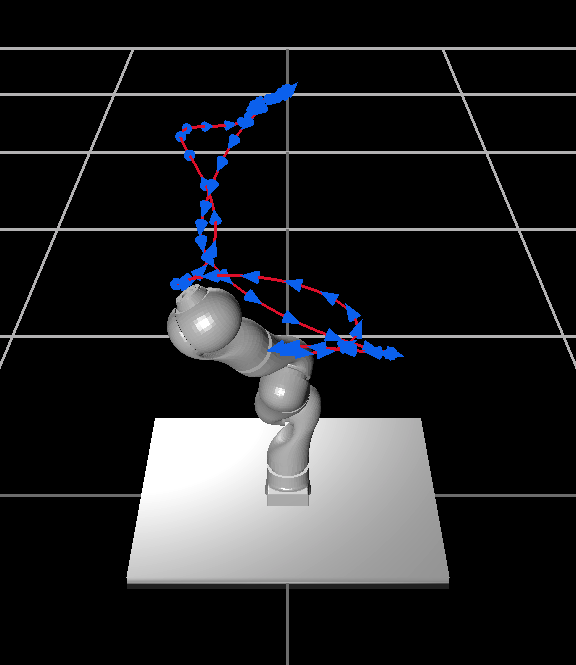}
      \includegraphics[width=0.2325\textwidth]{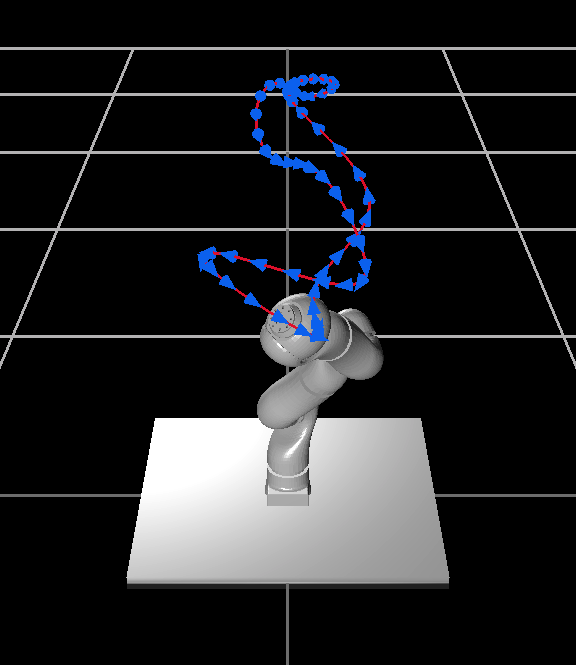}
      \includegraphics[width=0.2325\textwidth]{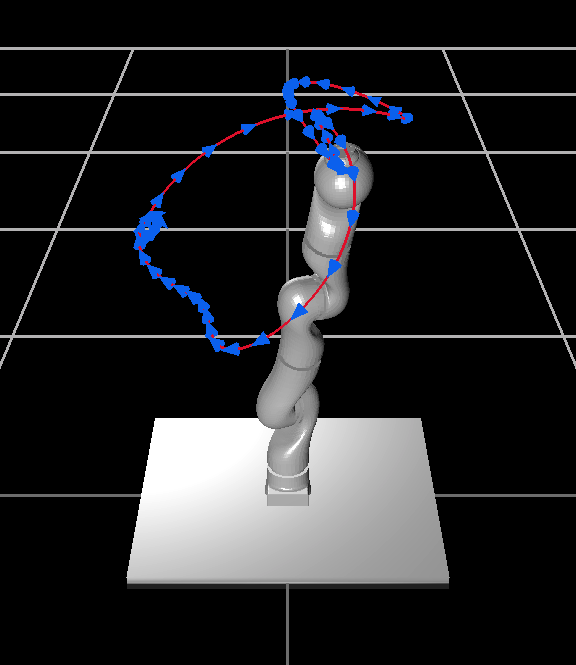}
      \includegraphics[width=0.2325\textwidth]{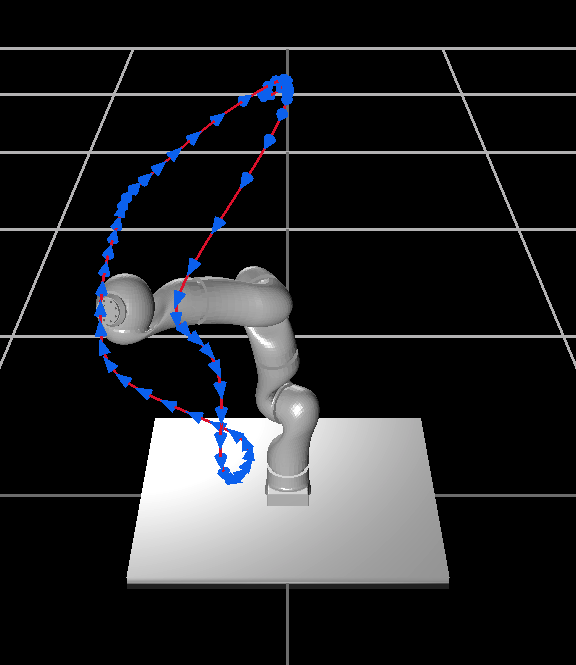}
      \vspace{0.1cm}
      \includegraphics[width=0.2325\textwidth]{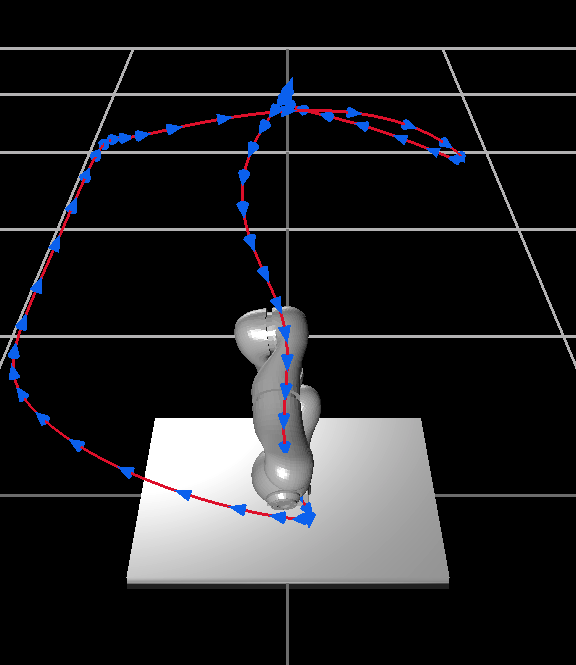} \includegraphics[width=0.2325\textwidth]{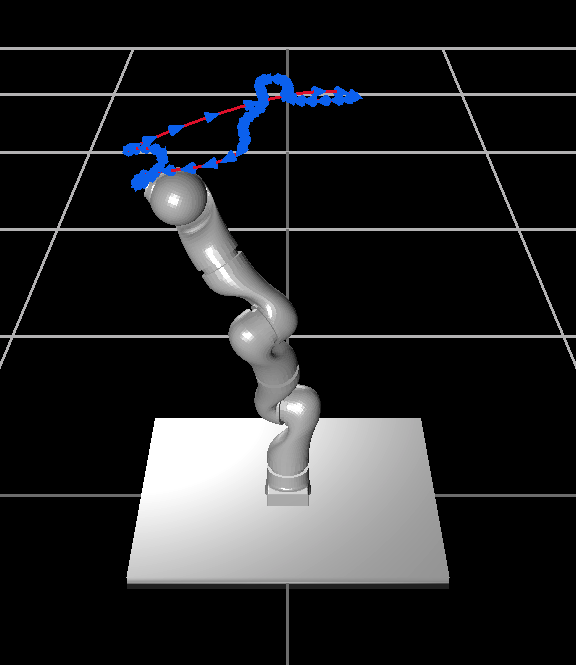}
      \includegraphics[width=0.2325\textwidth]{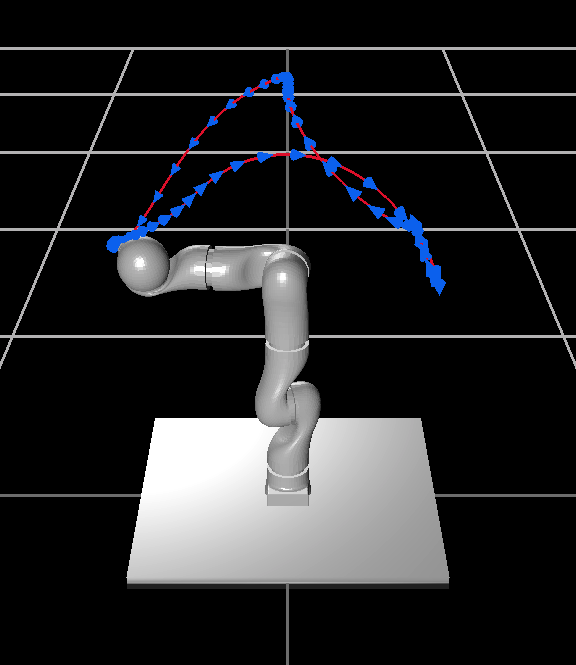}
      \includegraphics[width=0.2325\textwidth]{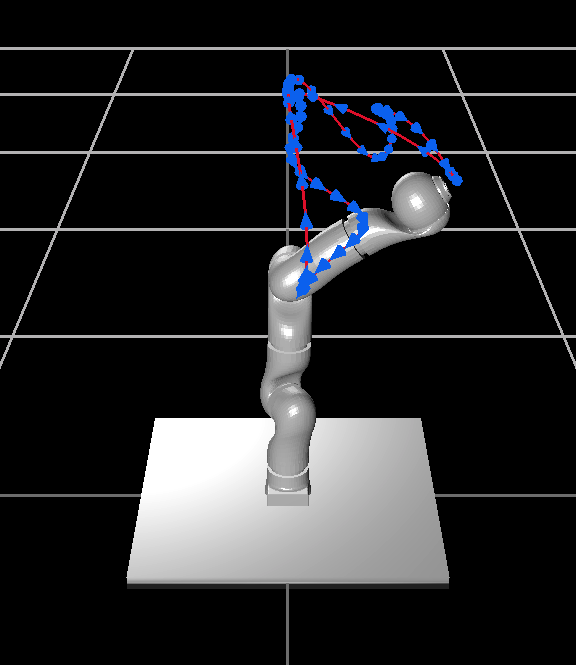}
    \end{minipage}\vspace{0.2cm}\begin{minipage}{1.03\columnwidth}
      \centering
      \includegraphics[width=0.305\textwidth]{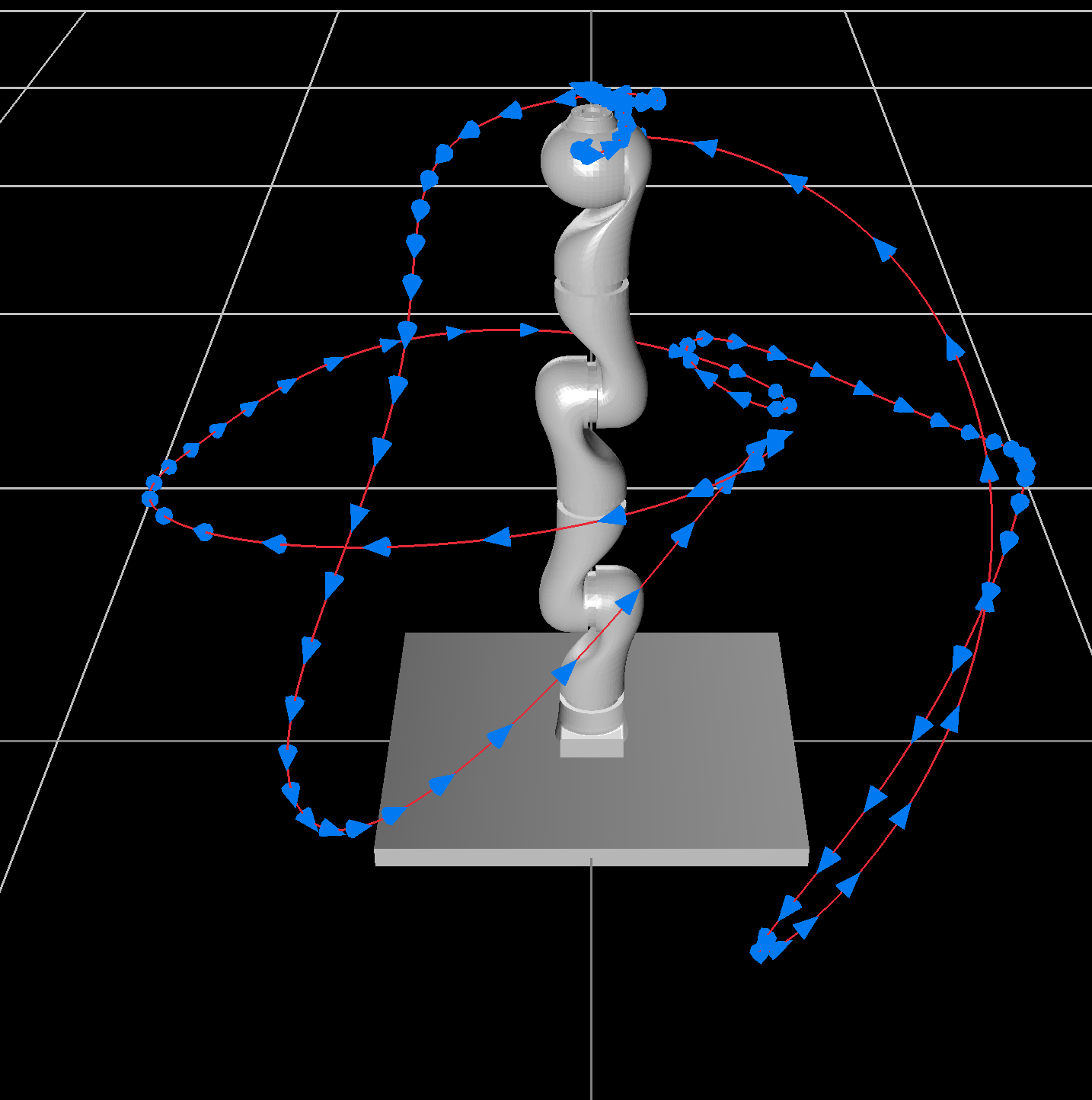}\includegraphics[width=0.305\textwidth]{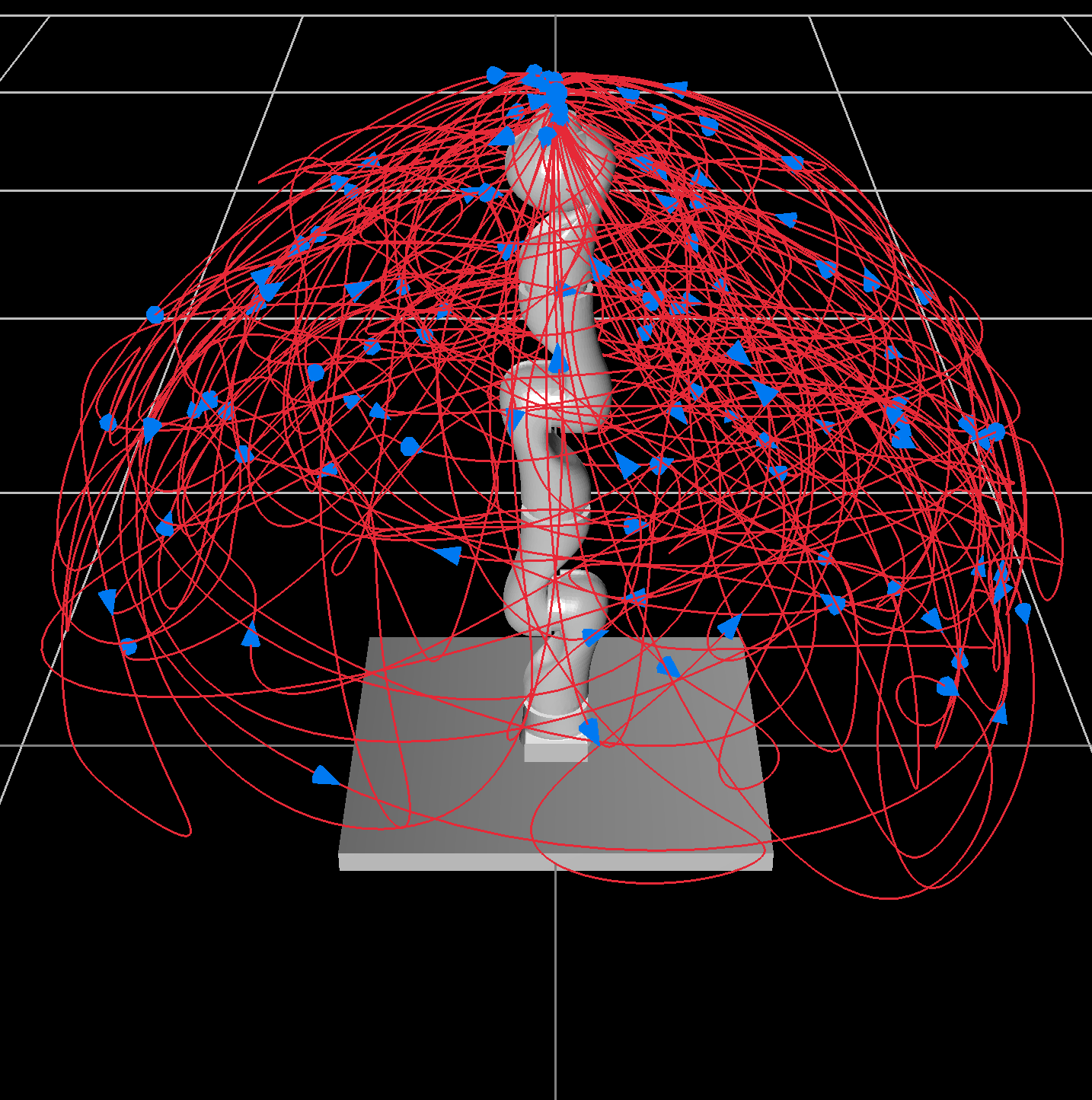}\includegraphics[width=0.305\textwidth]{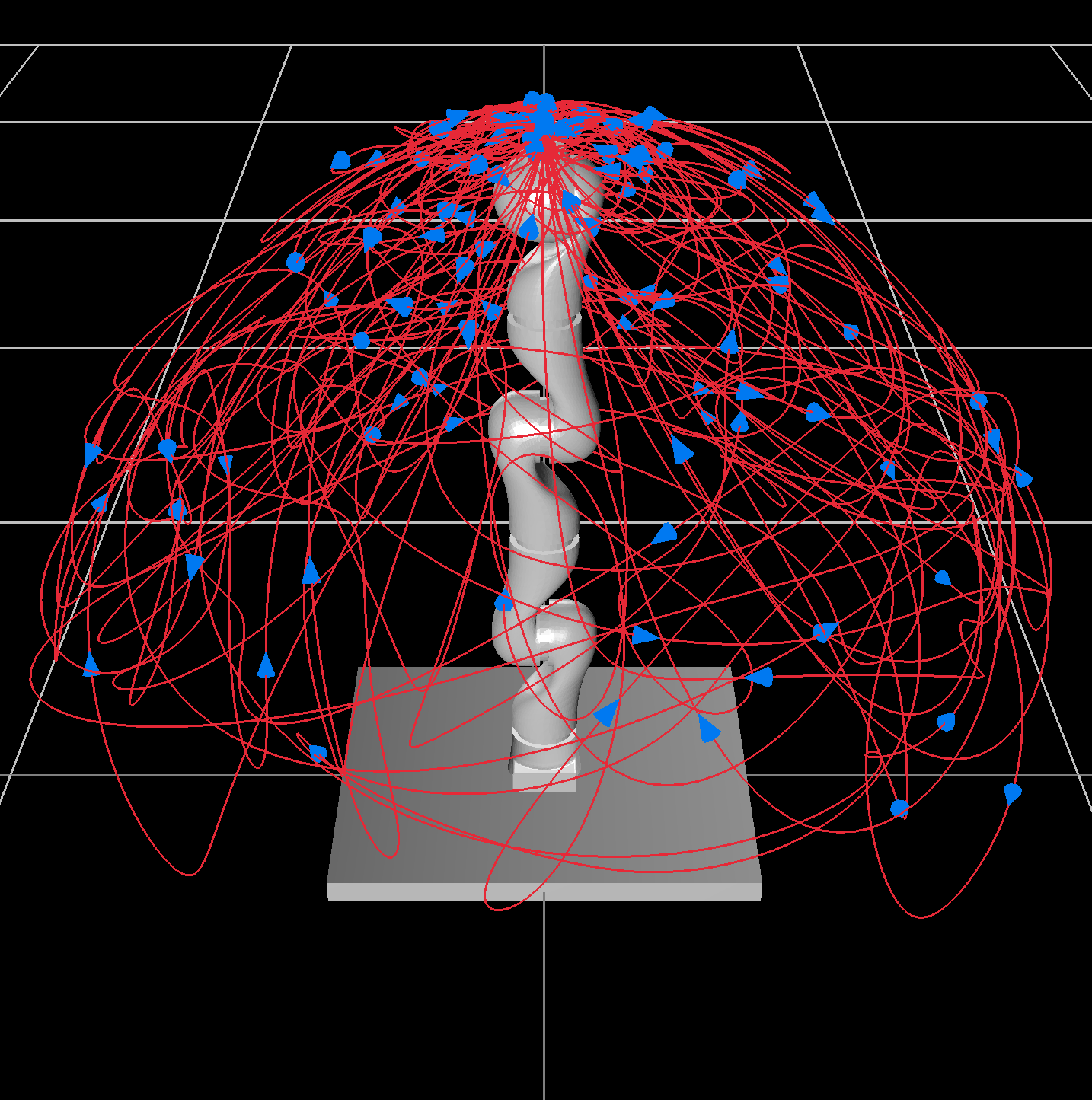}
      \includegraphics[width=0.1525\textwidth]{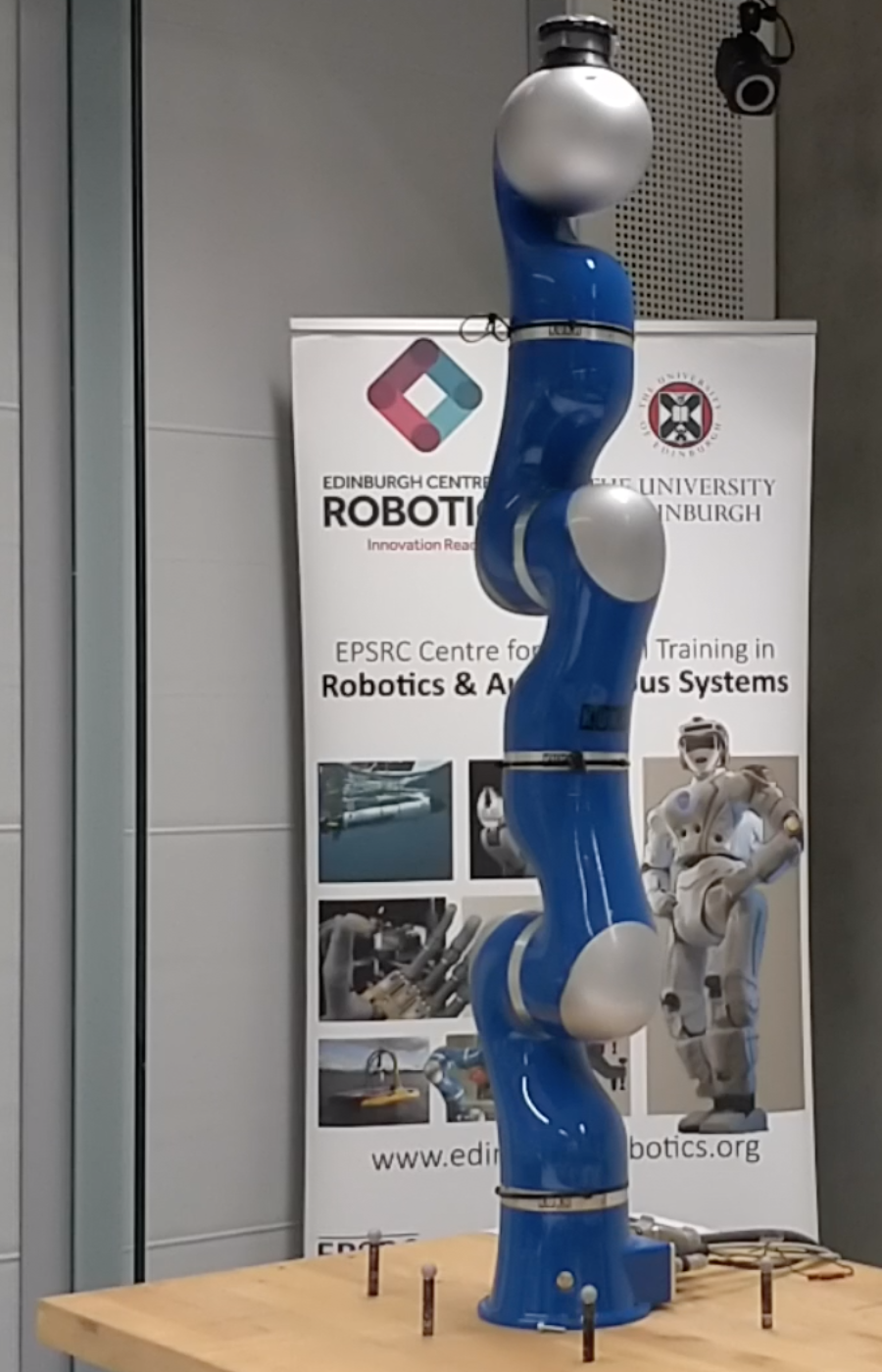}\includegraphics[width=0.1525\textwidth]{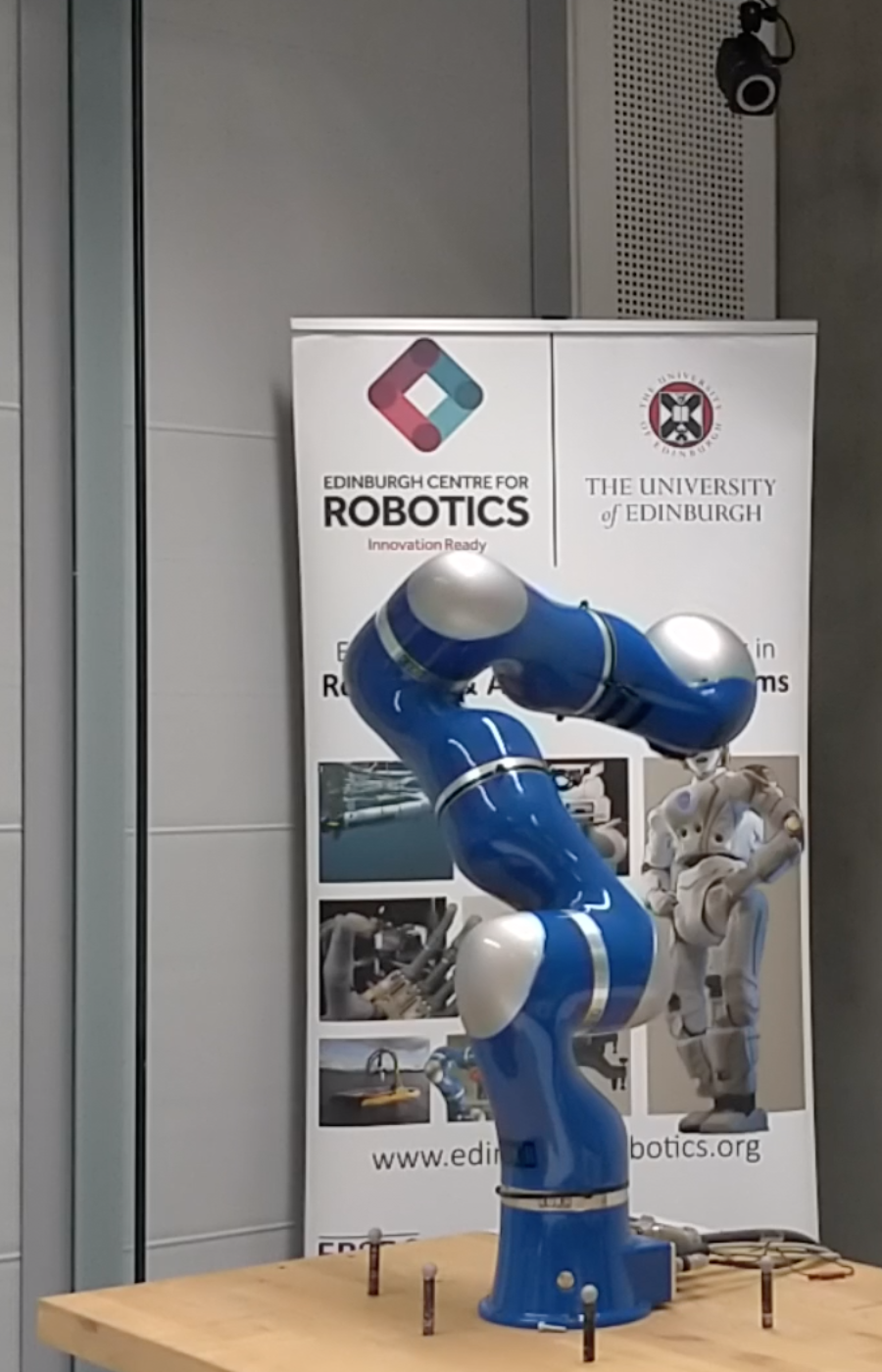}\includegraphics[width=0.1525\textwidth]{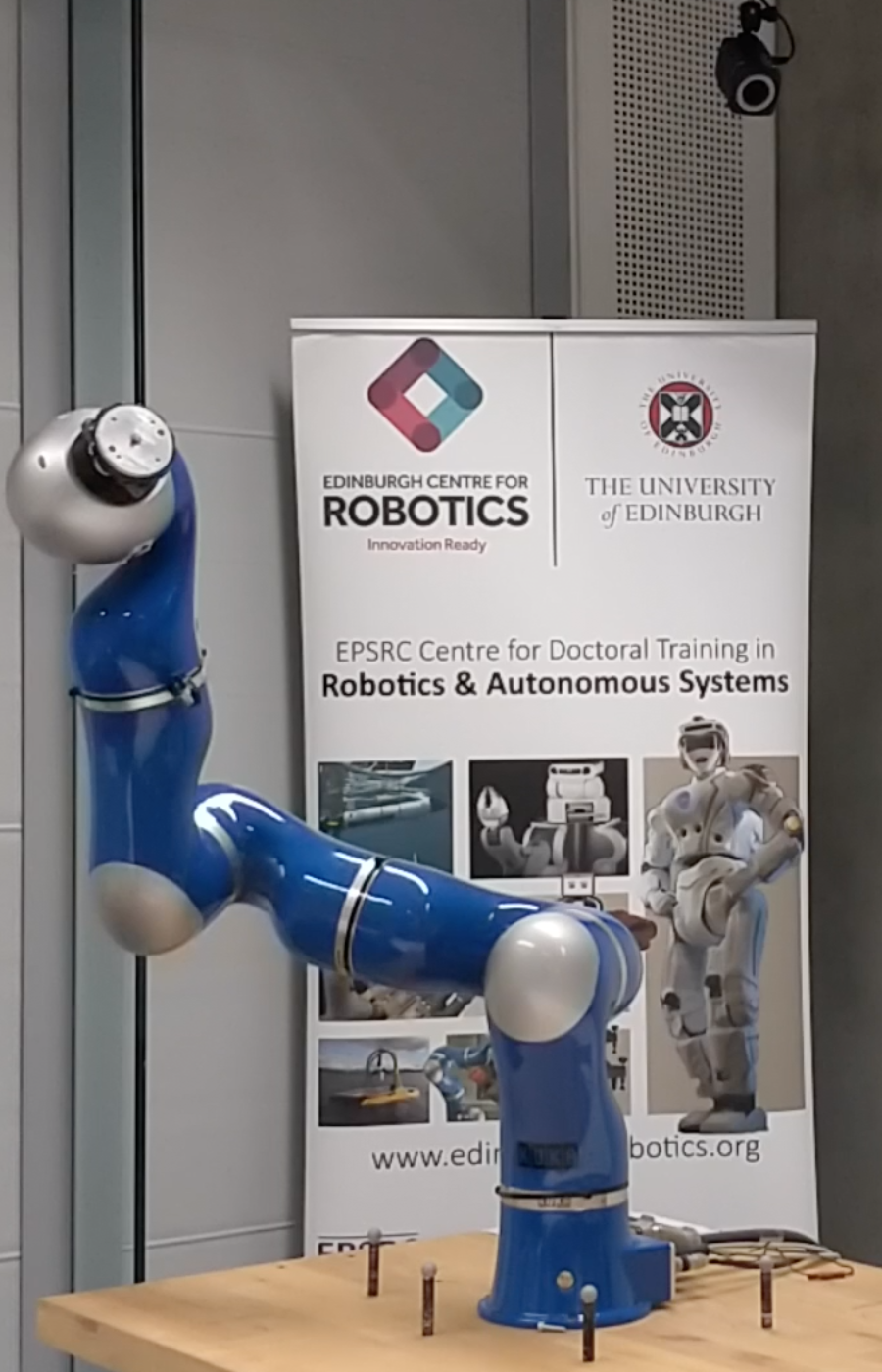}\includegraphics[width=0.1525\textwidth]{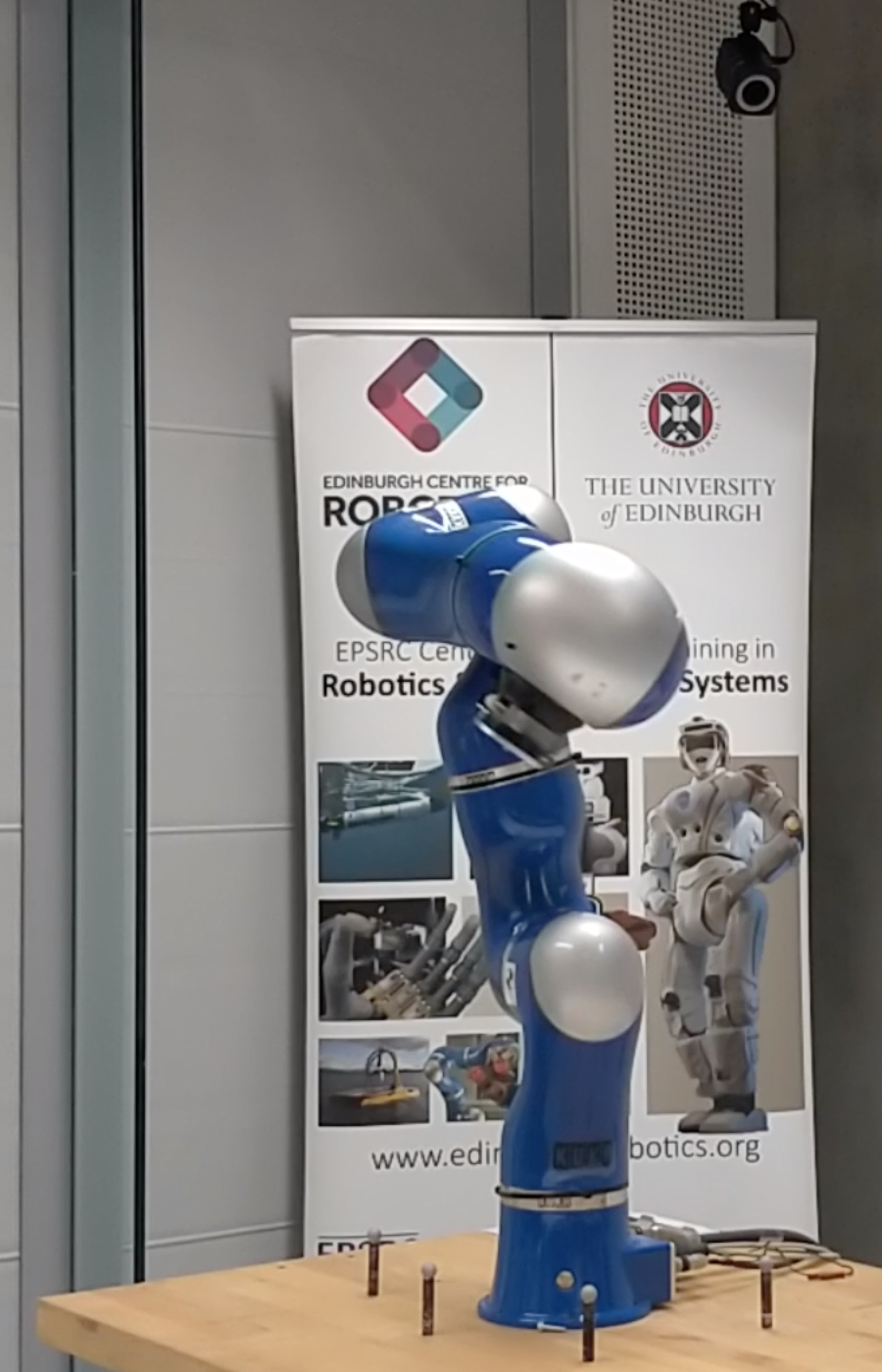}\includegraphics[width=0.1525\textwidth]{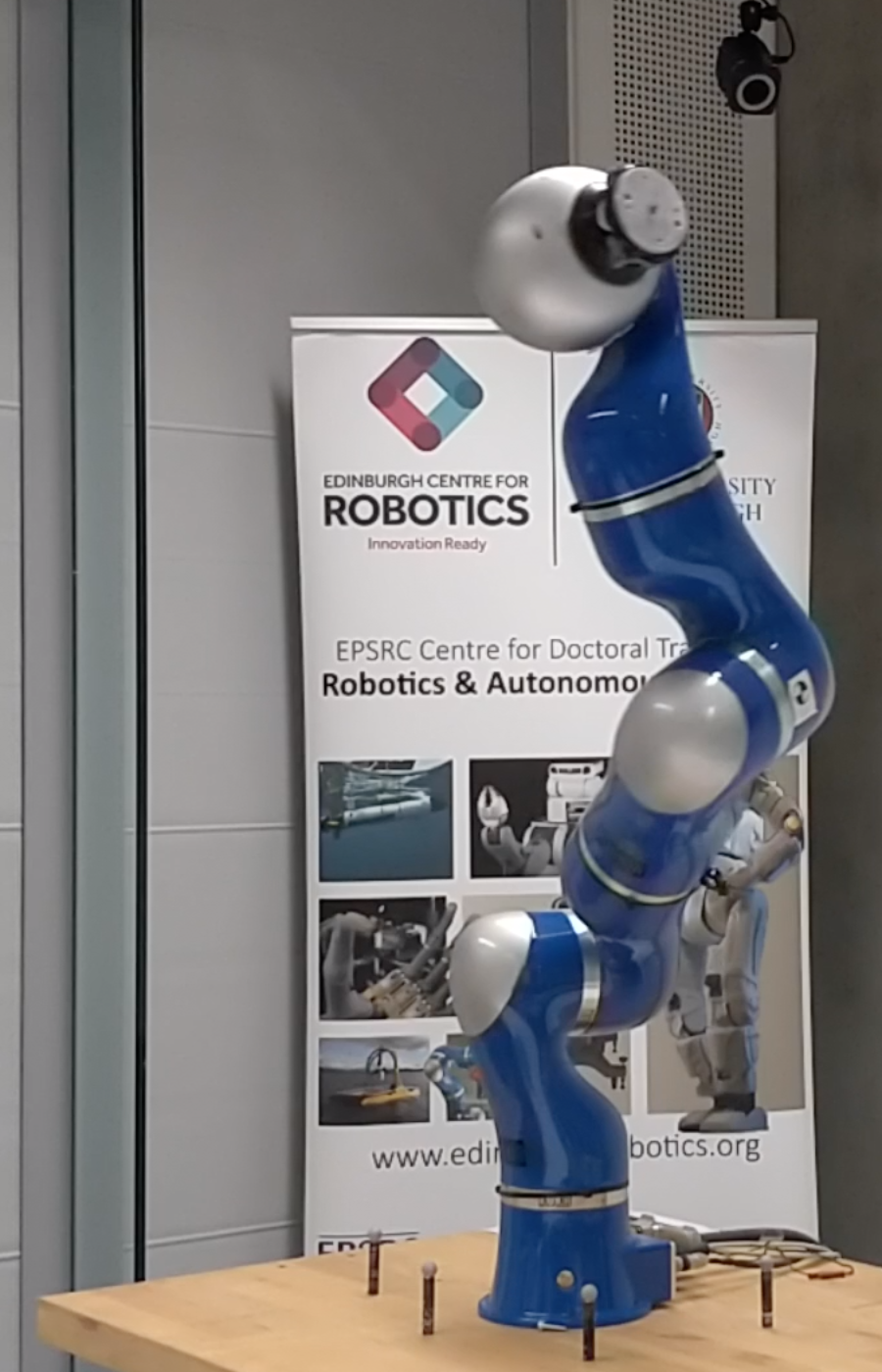}\includegraphics[width=0.1525\textwidth]{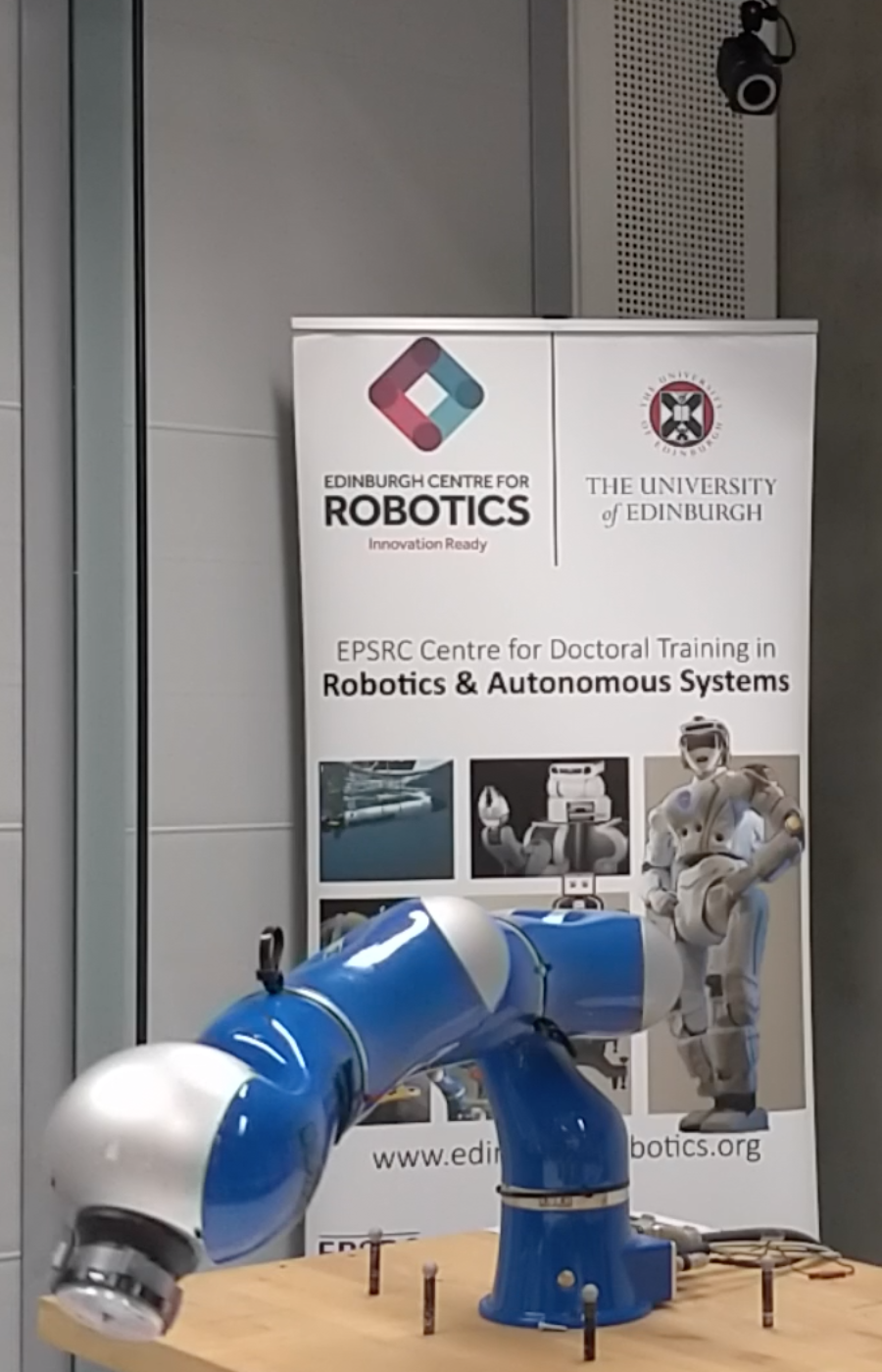}
    \end{minipage}
    \vspace{-6pt}
    \caption{\textbf{Visual results: Left:} examples of the SIDE-GAN generated trajectories - red traces show the end-effector positions. Trajectories are clearly spatially diverse. \textbf{Top-right:} examples of the single 560$s$ trajectory vs. the original seed set and the SIDE-GANs $35\times16s$ trajectories stacked together. \textbf{Bottom-right:} a few snapshots of one of the 16$s$ SIDE-GAN trajectories.}
    \label{figure:trajectories} 
    \vspace{-5pt}
\end{figure*}


We build our model on the top of a typical Deep Convolutional Generative Adversarial Network architecture, i.e., DCGANs \cite{DCGANs}\changed{, all nets have 4 layers, RELU-activated, using 100D uniform random noise vectors}. While conventional DCGANs generate images, we modify them to generate $7\times6\times2$ tensors representing Fourier transform parameters, which define short cyclic trajectories. For our goal of system identification, there are a number of extensions required to adapt DCGAN to generate trajectories that are valid (e.g. non-colliding) and diverse in both control parameter and regressor eigenvalue (inertial parameter) space. These are detailed as follows:

\noindent\textbf{Success Predictor Loss:} Vanilla GAN does not ensure that the majority of generations are valid (i.e., non-self colliding, or constraint-violating). To address this, we define a success predictor, as a shallow convolutional network mapping generated trajectories to a valid/faulty label. We pre-train this network to differentiate the initial dataset of valid trajectories, and some invalid trajectories generated by vanilla GAN. The trained success predictor has 99\% validation accuracy. We fix its weights and use it as a loss for the main SIDE-GAN training, thus encouraging it to generate valid trajectories. 

\noindent\textbf{Trajectory-to-eigenvalues converter} The salient feature space for analysis of trajectories is the eigenvalues of the rolled-out trajectory. To predict these features for generated trajectories, we pre-train a shallow convolutional network to map fourier trajectory parameters to the resulting eigenvalues. As above, we freeze its weights before plugging it into SIDE-GAN. The estimated eigenvalues are then used by the following two modules: 

\noindent\textbf{Eigenvalue Discriminator}. The basic GAN discriminator differentiates real vs fake trajectories in the GAN's  raw output space (modified Fourier parameter tensors). However, for our purposes, the crucial property of the trajectories is to cover the eigenvalue space well. Therefore we define a second discriminator that differentiates real/fake samples based on the eigenvalues of the rolled out trajectories -- as predicted by the eigenvalue estimator defined above. This is learned jointly with the vanilla discriminator in SIDE-GAN.

\noindent\textbf{Fitness Loss}. {SIDE-GAN so far aims to generate trajectories that are valid, and indistinguishable from the seed set used for training. Nevertheless, other things being equal, for SIDE purposes, we prefer trajectories with a more uniform eigenvalue distributions. We therefore define a final loss that penalizes the eigenvalue fitness (Sec III). This is trained jointly with the other SIDE-GAN modules, but activated after epoch 10 once the rest of the model has stabilized.}

\noindent\textbf{Alternative Diagonal Architecture}. The above three modules (converter, second discriminator, fitness loss) are based on eigenvalues. We also compare an alternative approach based on diagonal Fitness (Sec III).  In this case the converter estimates the $Y^TY$ diagonal, the discriminator discriminates based on this diagonal, and the fitness is defined as in Eq.~\ref{eq:diagFitness}.

\keypoint{Training SIDE-GAN} the generator produces a batch of trajectories defined by `fake' modified Fourier transform parameters. These are mixed with the real trajectories, and the first discriminator labels these are real or fake, and the success predictor labels them as valid or faulty. The converter  translates trajectories into pseudo-eigenvalues, which are then used by the second discriminator for labelling as `real' or `fake'. Finally, the pseudo-eigenvalues are also used to calculate and penalize the eigenvalue fitness metric. \cut{We use two generator iterations for each discriminator iteration, which we empirically found led to am ore stable loss convergence.}
{The training objective of SIDE-GAN is to produce diverse trajectories with small fitness loss and good validity scores.}

The full training objective has the following form:
\begin{equation}
\begin{aligned}
\underset{G}{\textrm{min}}\,\underset{D}{\textrm{max}}\,V(G,D)=\mathbb{E}_{\mathbf{x}\sim p_{data}(\mathbf{x})}[\log D(\mathbf{x})]\\
+\mathbb{E}_{\mathbf{z}\sim p(\mathbf{z})}[\log(1-D(G(\mathbf{z}))))]\\
+\mathbb{E}_{\mathbf{z}\sim p(\mathbf{z})}[P_{SR}(G(\mathbf{z})) + F(G(\mathbf{z}))]
\end{aligned}
\end{equation}

where $\mathbf{x}$ stands for a data example, $\mathbf{z}$ a random noise vector, $G(\mathbf{z})$ is a sample from the generator, $D(\mathbf{x)}$ represents the discriminator's estimate of the probability that $\mathbf{x}$ is from the real data set rather than from the generator, and $D(G(\mathbf{z}))$ - a probability that data are generated. $P_{SR}(G(\mathbf{z}))$ represents the predicted success rates of the generator output (i.e., the proportion of valid trajectories amongst the generated data). $F(G(\mathbf{z}))$ is the predicted fitness of the generator output.

We use the seed set to train the SIDE-GAN for 50 epochs in total, using two generator iterations for each discriminator iteration, and then discard everything but the generator. 

\keypoint{Applying SIDE-GAN:} The trained generator network provides our surrogate model for trajectory generation. It can generate as many diverse, new and exciting trajectories as necessary. Since it generates optimization-free, in a single forward pass, it can produce novel trajectories near instantaneously compared to traditional optimisation. 

\section{Experiments}
\subsection{Training Data \& Metrics}\label{training_data}
The data we use for training SIDE-GAN were acquired using a genetic optimizer for a rough global solution (ant colony based \cite{ANT})
followed by a finite difference gradient-based solver to fine-tune the trajectory locally~\cite{compassSearch}, using the pagmo2 library \cite{pagmo2}. The dynamics and regressors for the task were computed through the ARDL library \cite{smith2020online}. The full seed training set consists of 1800 cyclic trajectories of 16 seconds each, that are represented in terms of $7\times6\times2$ tensors or parameters of the modified Fourier transform (these correspond to the number of manipulator joints, the number of Fourier transform parameters, and 2 points defining the cyclic trajectory). Corresponding 35 eigenvalues of the resulting trajectory are used during the training to optimise the quality of the trajectories in terms of inertial parameter exploration. 

All of the performance results in this paper are averaged across at least three complete trainings of SIDE-GANs from scratch, and 10 batches of trajectories produced from different noise vectors.

\subsection{SIDE-GAN Training} We first answer the question: `\emph{Can a neural network learn to generate valid, exciting,  novel, and diverse trajectories?} To answer this, we analyse SIDE-GAN training dynamics in terms of success rate\footnote{Ratio of valid generated trajectories. I.e. non-self-colliding, obeying the joint velocity and position constraints, as well as the physical space constraints (e.g. avoiding the area within 10 cm above the desk on which the manipulator is set up).}, fitness (as defined in Sec~\ref{sec:motivation}), and the diversity of trajectories within a generated batch. Generation of trajectories that exhibit diversity between themselves is necessary, because if a trajectory generator simply repeats itself, little new information will be added when short trajectories are combined into a longer one for execution. For diversity, we use two metrics: (1) Average euclidean distance between raw trajectory Fourier parameters, (2) Average euclidean distance between the  $\text{diag}(Y^TY)$ vector of each trajectory. These metrics thus cover  diversity in both spatial and inertial parameter perspectives.
    
Figure~\ref{figure:gan_training} shows the training dynamics of SIDE-GAN using the 1800 seed trajectories as training data, and compares vanilla DCGAN with our two (eigenvalue, and diagonal-fitness) variants. \changed{Given the DCGAN-based architecture and the diversity of the training set, we do not expect any of them to mode-collapse.} From the plots we can see that training dynamics are somewhat unstable as per-usual with GANs: (1) Both SIDE-GAN variants quickly learn to reliably generate successful trajectories, while vanilla DCGAN struggles to pass 40\% success rate. (2) Vanilla DCGAN initially generates better fitness than SIDE-GAN, but by later epochs (40-50) SIDE-GAN produces better fitness. (3) In terms of the batch diversity, after 40 training epochs SIDE-GANs have greater spatial diversity than vanilla DCGAN and comparable overall spacial diversity to the original seed set. In inertial parameter space SIDE-GANs are significantly more diverse than the original seed set and usually equal or better than vanilla DCGAN. We \changed{expand the initial limited dataset by generating} 4200 novel trajectories \cut{from}\changed{by} SIDE-GAN after 50 \changed{training} epochs, and\cut{ together with the initial data,} this \cut{provides}\changed{expansion results in} a total set of 6000 short trajectories \cut{which are} used in the following system identification experiments. Figure~\ref{figure:trajectories} visualizes a few sample trajectories from SIDE-GAN. This visualisation shows \changed{considerable diversity}\cut{that they are very diverse}, at least in trajectory parameter space.

We also tried training some conventional non-neural network generative models such as Kernel Density Estimatiors (KDE)s. However, lacking a discriminator to provide a strong objective, these completely failed to produce valid trajectories, with an overall success rate under $0.1\%$.


\subsection{Exp 1: Multi- vs Single-Trajectory System Identification}

We first evaluate our idea of generating a set of smaller trajectories against standard practice of optimising the longest single trajectory that is computationally feasible. We stress that the key fundamental advantage our approach is: (1) Scalability to effectively unlimited total length, and thus much greater total excitation, unlike single trajectories. (2) Enabling massive parallelization for fast generation at any scale. Nevertheless, for the purpose of this experiment, we put these points aside and focus on comparing identification performance using a \emph{fixed} total length -- generated by a single trajectory optimisation, or our multi-trajectory optimisation. 

\keypoint{Setup: Dataset} To compare the trajectory optimisation methods, controlling for trajectory length, we optimise a single long trajectory of 560 seconds (the longest we can feasibly optimise) using eigenvalue-fitness criterion, and compare it to the concatenation of 35 short 16-second cyclic trajectories. As discussed earlier, our full dataset is 1800+4200=6000 trajectories. Thus we define a greedy strategy to pick a good subset 35 trajectories for direct comparison.
To achieve this, we first compute the diagonal of $Y^{T}Y$ for each trajectory, which we shall denote by $\psi_{i}$ for the $i$th trajectory. Each trajectory is then scored the following metric $d_i$ that balances preference for batch-diverse and exciting trajectories:
\begin{gather}
    g_{i} = ||\psi_{i} - \psi_{\text{previous best}}||\\
    f_{i} = \text{sum}(\psi_{i})\\
    d_{i} = \frac{g_{i}}{\text{max}(g)} +\frac{f_{i}}{\text{max}(f)}
\end{gather}

We greedily pick the trajectory with greatest score $d_i$, 
where `previous best' is the previously selected best trajectory, and initially $ \psi_{\text{previous best}}$ is set to zero. Thus we prefer those that are far from the previous choice, and have large diagonals. After each trajectory is selected, it gets removed from the set of trajectories to choose from next.



\keypoint{Setup: System Identification} We next use recursive least squares (RLS) to perform system identification and learn the dynamics of a Kuka LWR IV platform using the conventional long, conventional short (35 of 1800), and extended SIDE-GAN (35 of 6000) generated trajectories.
We use the fitted model to perform torque prediction and report the torque prediction accuracy (normalized mean squared error, nMSE) in Table~\ref{results_long_vs_short}. We repeat this experiment using both the simulated platform (perfect dynamic model with 10\% uniform random noise) via ARDL library \cite{smith2020online} and a real robotic arm.


\keypoint{Results}
The results show that, controlling for trajectory length: (1) Our multiple trajectory approaches clearly outperform the conventional single long trajectory approach both in simulation and on the real KUKA LWR. (2) The additional excitation trajectories synthesised by SIDE-GAN produce a small improvement over multiple trajectory optimisation. 

In terms of compute requirements: The single 560$s$ large trajectory generation required 14 hours, the 1800 seed trajectories ($\approx8h$ length) required 30 hours to generate (parallelised), the SIDE-GAN required a further 40 minutes to train, but thereafter can generate short 16$s$ trajectories in 1.4$ms$ per trajectory per thread, compared to 6 minutes per short trajectory using the conventional optimisation. 

Overall, we conclude that multi-trajectory optimisation performs favorably compared to the conventional approach, and especially with SIDE-GAN can easily be scaled to generating more excitation trajectory data for system identification. In the next section, we explore the benefit of using the full generated dataset for dynamics learning.

\cut{First, nMSE results for torque prediction task are presented in Table~\ref{results_long_vs_short}. They are comparing the sets of 35 multiple shorter 16 seconds trajectories (selected by greedy search from the original training set) versus the single conventional trajectories of corresponding length ($16\times35$ seconds). These prove multiple trajectory approach for the system identification beats the conventional single trajectory approach for the trajectories of similar cumulative length.}
\cut{And finally, we show that these results hold on the real robot. More specifically we re-run the very first experiment discussed in this section on KUKA LWR IV platform.}

\setlength{\tabcolsep}{0.33em}
\begin{table}

\vspace{0.2cm}
\begin{center}
\ra{1.3}
\begin{tabular}{c c c c c }
\hline
Trajectories:  & Single Long & Multiple original & Multiple SIDE-GAN\\
\hline
\hline
\multicolumn{1}{l}{Simulator} & $0.2248 \scriptsize (\pm 0.056)$ & $0.0063 \scriptsize (\pm 0.002)$ & $0.0028 \scriptsize (\pm 0.001)$ \\
\multicolumn{1}{l}{Real Robot} & $8.509 (\pm 9.906)$ &  $0.0239 (\pm 0.017)$ & $0.0210 (\pm 0.016)$\\
\hline
\end{tabular}
\end{center} \vspace{-5pt}
\caption{\textbf{Average nMSE across joints.} Multiple (35) short cyclic trajectories show better performance than the conventional single longer trajectory of comparable length. The best 35 short trajectories generated with SIDE-GAN further improve nMSE over those from the original seed set.
}
\vspace{-7pt}
\label{results_long_vs_short}
\end{table}

\subsection{Exp 2: System Identification with SIDE-GAN}\label{sec:exp2} We next evaluate system identification performance when using our full SIDE-GAN generated dataset. 

\keypoint{Torque Prediction} We first evaluate torque prediction, as in the previous experiment. We compare the margin of improvement between: (1) Seed+GAN generated data, using several GAN variants and (2) The  original seed data alone ($\times1$), and (3) The seed data, replicated $\times4$ or $\times10$ times (each replication is done with 10\% uniform random noise on both the robot state $(q,\dot{q},\ddot{q})$ and on the output torques). The $\times4$ replication corresponds to a similar amount of data to our GAN-generated dataset, and the $\times10$ replication corresponds to significantly more data than our GAN-generated dataset. 

The results in Table~\ref{results_torque_predictions} are reported in terms of \% improvement in torque prediction nMSE. We can see that: (1) Vanilla DCGAN already leads to a clear improvement, and (2) Our two SIDE-GAN variants further improve on vanilla DCGAN trajectory generation, (3) Comparing our two SIDE-GAN variants that use eigenvalue or diagonal-based fitness, the eigenvalue-fitness variant performs best. (4) Simply replicating the original seed data does provide a simple alternative: the margin of our methods over the original data results does not decrease systematically with replication factor. 

\keypoint{Parameter Estimation Quality} We next investigate the parameter estimation quality for the different methods. We quantify estimation quality by the norms of the diagonal of the covariance from the RLS algorithm, which gives an estimate of the uncertainty of the internal base parameters. 

The results in Table~\ref{results_covariance_norms} shows that SIDE-GAN eigenvalue-fitness variant provides the lowest (least uncertain) norm estimates compared to both the original data and any of the other competitors.

\setlength{\tabcolsep}{0.31em}
\begin{table}

\vspace{0.2cm}
\begin{center}
\ra{1.3}
\begin{tabular}{c c c c c }
\hline
Original + generated data vs. & orig. ($\times 1$)&  orig. ($\times 4$)& orig. ($\times$10)\\
\hline
\hline
\multicolumn{1}{l}{+ Vanilla DCGAN} & $36\% \scriptsize (\pm 44\%)$ & $21\% \scriptsize (\pm 30\%)$ & $38\% \scriptsize (\pm 17\%)$\\
\multicolumn{1}{l}{+ SIDE-GAN ($Y^{T}Y$ eigens)} & $\textbf{51\%} \scriptsize (\pm 27\%)$ & $\textbf{39\%} \scriptsize (\pm 25\%)$ & $\textbf{53\%} \scriptsize (\pm 8\%)$\\
\multicolumn{1}{l}{+ SIDE-GAN ($Y^{T}Y$ diag.)} & $\textbf{52\%} \scriptsize (\pm 16\%)$ & $35\% \scriptsize (\pm 32\%)$ & $44\% \scriptsize (\pm 27\%)$ \\
\hline
\end{tabular}
\end{center}
\vspace{-5pt}
\caption{Improvements in nMSE of torque predictions with respect to the original training data (fed into RLS $\times 1$, $\times 4$, and $\times 10$ times with 10\% uniform random noise). Improvement ratio is averaged across the joints. 
}
\label{results_torque_predictions}
\end{table}

\begin{table}

\vspace{0.2cm}
\begin{center}
\ra{1.3}
\begin{tabular}{c c c c c }
\hline
Methods  & Covariance Diagonals Norms \\
\hline
\hline
\multicolumn{1}{l}{Original dataset $\times 1$} & 2.65384  \\
\multicolumn{1}{l}{Original dataset $\times 4$} & 2.44338 \\
\multicolumn{1}{l}{Vanilla DCGAN} & 2.2573  \\
\multicolumn{1}{l}{SIDE-GAN ($Y^{T}Y$ eigens)} & \textbf{1.89143}  \\
\multicolumn{1}{l}{SIDE-GAN ($Y^{T}Y$ diag.)} & 2.36345 \\
\hline
\end{tabular}
\end{center}
\vspace{-5pt}
\caption{Parameter estimation quality. Lower values mean that  covariance matrices have smaller determinants, which means inertial parameters are predicted with more certainty.  
}
\label{results_covariance_norms}
\end{table}

\subsection{Discussion}
\keypoint{SIDE-GAN Dependence on Seed Set Size}
Our experiments used a fixed seed set of 1800 trajectories throughout to train SIDE-GAN. We explored reducing the training set size. Training on, e.g., 240 trajectories, SIDE-GAN still generates diverse and exciting trajectories. However, this does not provide sufficient data for the GAN to learn the constraints well, and the validity rate of generated trajectories suffer (about $10\%$). This could still be useful since many samples can be drawn and invalid trajectories filtered. Generating and filtering in this way is still faster than conventional optimization (which takes roughly 6 minutes per trajectory) vs SIDE-GAN (32 per second, including checking the validity).


\keypoint{SIDE-GAN Generality}
GAN-based methods are generally indifferent to the specifics of the training data. Thus the SIDE-GAN method is expected to work well for other types of manipulators, and other dynamics models. That said, SIDE-GAN does need a seed set for the relevant manipulator. The trajectories generated by the SIDE-GANs are unlikely to generalise effectively across manipulators. 


\section{Conclusions and Future Work}
This work shows the benefits of using multiple trajectories instead of the conventional single parameterised trajectory for the task of the system identification and torque prediction. 
Further, it proposes a method for generating valid and more diverse trajectories for the above task at the speed exceeding the underlying method by at least two orders of magnitude. The trajectory generator is trained to produce diversity in both trajectory and inertial parameter space. Numerical results on trajectory validity, fitness metrics, and torque prediction -- in both simulation and on real Kuka arm -- confirm our contributions.

In future work we intend to investigate the use of conditional models  to incorporate user-specified specific region-based and inertial-parameter-based focused exploration. We will also explore sparse data transfer learning to reduce the size of the seed set required to learn SIDE-GANs.





\bibliographystyle{IEEEtran}
\bibliography{IEEEexample}

\end{document}